\documentclass[runningheads]{llncs}
\usepackage{amsmath}%............................ Advanced math (before fonts)
\usepackage{amssymb}%............................ AMS Symbol fonts
\usepackage[hashEnumerators, smartEllipses]{markdown}
\usepackage{wrapfig}

\usepackage{parskip}%............................ Spacing Paragraphs
\usepackage[T1]{fontenc}%........................ Type 1 fonts
\usepackage{amsfonts}%............................AMS fonts
\usepackage{graphicx}
\usepackage{algcompatible}
\usepackage{algorithm}
\usepackage{caption} %many figures in one figure (note subfigure and subfig are deprecated)
\usepackage{subcaption} %many figures in one figure (note subfigure and subfig are deprecated)
\usepackage{algpseudocode,algorithmicx}
\usepackage{xcolor}
\floatname{algorithm}{Algorithm}
\algnewcommand{\algorithmicand}{\textbf{ and }}
\algnewcommand{\algorithmicor}{\textbf{ or }}
\algnewcommand{\algorithmicnot}{\textbf{ not }}
\algnewcommand{\algorithmicfalse}{\textbf{ false }}
\algnewcommand{\algorithmictrue}{\textbf{ true }}
\algnewcommand{\algorithmicbreak}{\textbf{break}}
\algnewcommand{\OR}{\algorithmicor}
\algnewcommand{\AND}{\algorithmicand}
\algnewcommand{\NOT}{\algorithmicnot}
\algnewcommand{\FALSE}{\algorithmicfalse}
\algnewcommand{\TRUE}{\algorithmictrue}
\algnewcommand{\Break}{\algorithmicbreak}
\newcommand{\compresslist}{ % Define a command to reduce spacing within itemize/enumerate environments, this is used right after \begin{itemize} or \begin{enumerate}
	\setlength{\itemsep}{1pt}
	\setlength{\parskip}{0pt}
	\setlength{\parsep}{0pt}
}
\algnewcommand\algorithmicclass{\textbf{class}}
\algdef{SE}[CLASS]{Class}{EndClass}[1]{\algorithmicclass\ \textproc{#1}}{\algorithmicend\ \algorithmicclass}

\algnewcommand\algorithmicenum{\textbf{enum}}
\algdef{SE}[ENUM]{Enum}{EndEnum}[1]{\algorithmicenum\ \textproc{#1}}{\algorithmicend\ \algorithmicenum}

\algnewcommand\algorithmicwhen{\textbf{when}}
\algdef{SE}[WHEN]{When}{EndWhen}[1]{\algorithmicwhen\ #1 \algorithmicdo}{\algorithmicend\ \algorithmicwhen}

\algnewcommand\algorithmicconstructor{\textbf{constructor}}
\algdef{SE}[CONSTRUCTOR]{Constructor}{EndConstructor}[1]{\algorithmicconstructor\ (#1)}{\algorithmicend\ \algorithmicconstructor}
\begin{document}
\title{A Multi-Criteria Automated MLOps Pipeline for Cost-Effective Cloud-Based Classifier Retraining in Response to Data Distribution Shifts\thanks{Supported by Optimall Research Lab.}}
\titlerunning{Multi-Criteria Automated MLOps Pipeline}
\author{Emmanuel K. Katalay\inst{1}  \and David O. Dimandja\inst{1}  \and
Jordan F. Masakuna\inst{1} \orcidID{0000-0001-6165-8402}}
\authorrunning{Emmanuel and David and Jordan}
\institute{Department of Mathematics and Computer Science, University of Kinshasa
\\
\email{
jordan.masakuna@unikin.ac.cd} (corresponding author)\\
\url{https://www.fimproso.org} }
\maketitle              % typeset the header of the contribution
\begin{abstract}
The performance of machine learning (ML) models often deteriorates when the underlying data distribution changes over time, a phenomenon known as data distribution drift. When this happens, ML models need to be retrained and redeployed. ML Operations (MLOps) is often manual, i.e., humans trigger the process of model retraining and redeployment. In this work, we present an automated MLOps pipeline designed to address neural network classifier retraining in response to significant data distribution changes.  Our MLOps pipeline employs multi-criteria statistical techniques to detect distribution shifts and triggers model updates only when necessary, ensuring computational efficiency and resource optimization. We demonstrate the effectiveness of our framework through experiments on several benchmark anomaly detection data sets, showing significant improvements in model accuracy and robustness compared to traditional retraining strategies.  Our work provides a foundation for deploying more reliable and adaptive ML systems in dynamic real-world settings, where data distribution changes are common.

\end{abstract}
\section{Introduction}\label{sec:introduction}
Machine learning (ML) models are often deployed in dynamic environments where the underlying data distribution may evolve over time, leading to performance degradation. This phenomenon, known as data distribution drift \cite{lu2018learning}, presents a significant challenge for maintaining the reliability and accuracy of models in production. In many real-world applications, models trained on historical data may no longer reflect the current patterns in incoming data, which can negatively impact decision-making processes. Addressing this issue requires efficient and timely retraining of models to adapt to these shifts in data characteristics.

In this paper, we propose an automated ML Operations (MLOps) \cite{kreuzberger2023machine} pipeline for model retraining that responds to significant data distribution changes in classification tasks. The pipeline leverages a combination of statistical metrics to monitor, detect, and react to shifts in data distributions, triggering retraining only when necessary. This approach minimizes the need for manual intervention, enabling a more streamlined, cost-effective solution for model maintenance.

Retraining machine learning models typically involves several resource-intensive phases, including hyperparameter optimization \cite{feurer2019hyperparameter}, cross-validation \cite{berrar2019cross}, and multiple rounds of model evaluation, all of which contribute to substantial computational overhead. This process, particularly when done repeatedly or unnecessarily, can lead to high cloud computing costs, which are especially burdensome in production environments where models need to be frequently updated. By automating the retraining pipeline and focusing on significant distribution changes, our method avoids redundant computations, reducing the consumption of cloud resources \cite{sterling2017high} and ultimately lowering operational costs.

The proposed pipeline not only addresses the technical challenge of detecting and responding to distribution drift but also provides a clear economic advantage. In practice, optimizing the retraining schedule can save companies substantial amounts of money by avoiding the costs of inefficient retraining cycles. With our solution, organizations can mitigate the risk of over-spending on unnecessary retraining processes, while ensuring that their models remain accurate and up-to-date. We demonstrate the effectiveness of our approach through extensive experiments on benchmark classification datasets, showing how it outperforms traditional retraining strategies in both performance and cost efficiency.

\textbf{Contributions}. The integration of diverse statistical tests into a unified Continuous Integration/Continuous Delivery (CI/CD) pipeline for automated ML model retraining presents two significant contributions to the field:
\begin{itemize}
    \item \textbf{Enhanced and systematic model monitoring}. By systematically combining metrics (e.g., Population Stability Index (PSI) and Kullback-Leibler (KL) Divergence) with rigorous statistical tests, the pipeline provides a robust mechanism for detecting both data and model drift. This approach pinpoints when and precisely why a model's performance has degraded, making the need for retraining auditable and verifiable. This moves beyond simple accuracy checks toward deep, diagnostic monitoring. 

\item \textbf{Scientifically robust methodology for automation}. The combination of these stability metrics and formal statistical tests (integrated into a comprehensive retraining automation framework) offers a fresh and scientifically robust methodology. This framework provides the ML community with a highly adaptable, verifiable, and principle-driven standard for automating the model lifecycle, ensuring that retraining decisions are based on measurable and statistically sound evidence.
\end{itemize}

%The rest of the paper is organised as follows:
%
%Section \ref{sec:related-work} discusses the state of the art for decentralised computation of closeness centrality distribution of networks.
%The new algorithm will be discussed in Section \ref{sec:description}.
%
%Section \ref{sec:results} discusses the results. 
%
%In Section \ref{sec:conclusion}, we conclude and propose further work.

\section{Background and related work}
\label{sec:related-work}

\subsection{Background}

ML models, particularly those based on neural networks, have become central to many real-world applications such as image classification and natural language processing. These models are trained on historical datasets and fine-tuned through backpropagation to learn the underlying patterns in the data. However, in practical settings, these patterns may evolve due to changes in data distribution over time, which can  degrade model performance. This phenomenon is often referred to as data distribution (or concept)  drift \cite{lu2018learning}.

\subsubsection{Data distribution drift.}
Data distribution drift occurs when the statistical properties of incoming data change over time, such as shifts in the mean or variance of features. Such changes can render the model’s predictions less accurate, as the model is no longer aligned with the current data distribution. In classification tasks, for instance, this could manifest as a decrease in accuracy or an increase in misclassifications. Detecting and addressing these shifts promptly is critical for maintaining robust and reliable ML systems.

Assuming that $P_{t}$ represents the joint probability distribution between the input
variable $x$ and the target variable $y$ at time $t$, then concept drift will occur if (\ref{eq:drift1}) holds
when a time $t_0$ turns to $t_1$ \cite{bayram2022concept}.
\begin{equation}
    \label{eq:drift1}
    \exists x: P_{t_0}(x, y) \neq P_{t_1}(x, y)\,\text{ or }\, \exists x: P_{t_0}(x) P_{t_0}(y|x) \neq P_{t_1}(x) P_{t_1}(y|x)\,.
\end{equation}

\subsubsection{Neural networks and their use in classification.}
Neural networks \cite{dongare2012introduction}, and particularly deep learning models, are widely used for classification tasks due to their ability to model complex patterns in high-dimensional data. They consist of multiple layers of interconnected neurons that process input data through nonlinear activation functions. The model’s parameters are optimized during training to minimize the loss function, typically using gradient-based methods like stochastic gradient descent (SGD). Despite their success, neural networks face challenges when it comes to adapting to changing data distributions. When the model’s training data no longer reflects the characteristics of new incoming data, performance degrades. Thus, retraining becomes necessary to restore performance.

\subsubsection{MLOps and model maintenance.}
MLOps  is an evolving discipline focused on automating and streamlining the end-to-end lifecycle of machine learning models, including development, deployment, monitoring, and maintenance. Effective MLOps practices enable efficient model retraining, version control, and continuous integration (CI) and deployment (CD). One key aspect of MLOps is automating model retraining in response to changes in data distribution, which can help organizations maintain the accuracy and reliability of models without constant manual intervention. The automation of this process is crucial in scaling machine learning systems across large-scale production environments where retraining cycles may be frequent and computationally expensive.

\subsubsection{Statistical metrics for drift detection.}
Various statistical techniques have been developed to detect distribution shifts. These include hypothesis testing (e.g., Kolmogorov-Smirnov test \cite{berger2014kolmogorov}), divergence measures (e.g., Kullback-Leibler divergence \cite{ji2020kullback}), and more advanced techniques such as the Population Stability Index (PSI) \cite{yurdakul2018statistical} and Maximum Mean Discrepancy (MMD) \cite{smola2006maximum}. These methods compare the distributions of the data over time and flag significant changes, enabling the detection of data drift. For classification tasks, metrics like accuracy, precision, recall, and F$_1$-score \cite{masakuna2023prior,masakuna2020active} are used to evaluate the model’s performance and detect when drift impacts the model's prediction quality. By integrating these drift detection metrics into an automated pipeline, it becomes possible to trigger retraining only when a significant drift is detected, reducing the need for constant retraining cycles.

\subsection{Literature review}
Several key approaches have been proposed in the literature to address model retraining in the face of distribution shifts, with various methods for detecting drift and automating the retraining process.
One of the earliest and most influential works in drift detection is \cite{bifet2007learning}. They propose the ADaptive WINdowing (ADWIN) algorithm, which dynamically adjusts the window of data used for training and tests for changes in distribution. While this method is effective in detecting changes, it relies on continuous retraining and does not provide an integrated solution for automated model maintenance in production environments. ADWIN does not address the computational costs associated with retraining.
Ditzler et al. \cite{ditzler2015learning} provide an overview of the concept of drift-aware learning, where models continuously adapt to drift by using ensemble methods. Their approach uses multiple classifiers and a mechanism to detect which models perform well under the current data distribution. While this method improves model robustness to drift, it still requires substantial computational resources, particularly in maintaining multiple classifiers and evaluating their performance.
Other approaches explore strategies to reduce the computational overhead of retraining by proposing selective retraining based on model performance metrics. Their approach relies on continuously monitoring model drift and only triggers retraining when performance drops below a predefined threshold. However, their method does not incorporate the concept of statistically significant distribution changes and may retrain models unnecessarily.

Sculley et al. \cite{sculley2015hidden} laid the groundwork for modern MLOps by proposing a system for automating the training, testing, and deployment of ML models. However, this pipeline is focused on the general automation of ML workflows, rather than addressing the specific problem of data drift detection and model retraining in response to distribution changes. 
Kreuzberger et al. \cite{kreuzberger2023machine} provide an extensive overview of MLOps practices and propose frameworks for automating the retraining process based on various triggers, including drift detection. While their work emphasizes automation, it often lacks detailed strategies for cost optimization in large-scale systems, which is a crucial aspect for industries dealing with massive datasets.
In a related approach, Louppe et al. \cite{louppe2017learning} discuss the use of drift-aware model retraining in the context of probabilistic modeling and emphasize the importance of adaptive retraining strategies to avoid overfitting to past data distributions. Although their approach offers valuable insights into adaptive learning, the proposed methods are not directly optimized for classification tasks or for integration into production pipelines.

More recent work has focused on the economic aspect of retraining in production environments. Retraining a ML model is necessary to maintain its accuracy as data evolves, but it is costly because it often requires processing the entire dataset. The challenge is to balance retraining too frequently, which incurs high computing costs, and retraining too infrequently, which leads to outdated models and reduced accuracy. \cite{mahadevan2024cost} introduces the Cost-Aware Retraining Algorithm (CARA), which optimizes the decision of when to retrain by considering data, model performance, and query costs. Through experiments on both synthetic and real-world datasets, CARA outperforms drift detection baselines by achieving better accuracy with fewer retraining decisions.
Feurer et al. \cite{feurer2019hyperparameter}  overview methods for hyperparameter optimization, which is a critical part of retraining but often computationally expensive. While this is an important contribution, it  does not address the question of when to retrain a model, an essential aspect of cost-effective MLOps.

Our approach incorporates metrics for drift detection into a fully automated MLOps pipeline. This pipeline optimizes the retraining schedule, ensuring that models are only retrained when significant drift is detected, thereby reducing unnecessary  cloud resource usage. Compared to previous works, our approach is a MLOps automation of CARA but with rigourous data drift detection mechanism.

%Here some potential research challenges.
%\begin{itemize}
%\compresslist
%    \item  Choosing optimal thresholds for each metric: deciding when a drift is "significant enough" for retraining could require experimentation and domain knowledge.
 %  \item Balancing false positives and false negatives: Fine-tuning the metrics to avoid retraining too frequently or missing important drifts.
%   \item Data and concept drift detection in complex models: models like deep learning may have complex relationships that are harder to track through simple statistical tests.
%\end{itemize}

\section{Auto-MLOps Pipeline}
\label{sec:description}
The MLOps pipeline continuously monitors incoming data for distribution shifts, leveraging drift detection techniques. When a shift is detected, our pipeline performs an automated decision-making process that balanced model performance improvement against computational cost before triggering retraining.
\begin{figure}[h]
	\centering
	\includegraphics[width=1\textwidth]{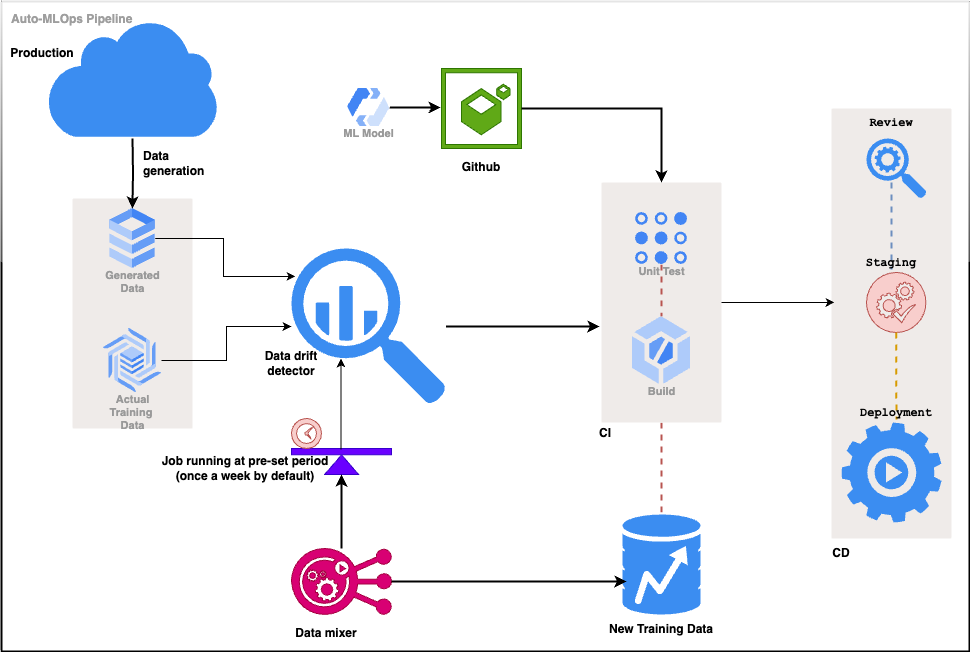}
	\caption{CI/CD pipeline for ML model deployment when data drift occurs.}
	\label{fig:pipeline}
\end{figure}
\subsection{Architecture}
In Figure \ref{fig:pipeline}, we show the automated CI/CD pipeline designed for building, testing, and deploying an ML model. Here is the role of each component:
\begin{itemize}
\compresslist
    \item \textbf{Production}. Data is generated from production environments. This could be raw user interaction data, logs, or sensor data, depending on the use case. 
    \item \textbf{Actual training data}. It is an historical or pre-processed data used as the baseline for training  ML models.
    \item \textbf{Data drift detector}.  
It continuously monitors and compares the newly generated data to the existing training data to identify ``data drift``.  
It alerts when data drift is detected and triggers a retraining pipeline if needed.
\item \textbf{Data mixer}. 
When receives a signal from data drift detector, it combines the newly generated data with the existing training data to create a ``new dataset`` for model training.
%\item \textbf{GitHub}. It stores the source code, including the ML model's training, testing, and deployment scripts. 
\item  \textbf{CI}.   
It validates the ML code, ensuring changes do not introduce bugs or issues (i.e., unit test). These tests include model validation and performance benchmarks.
It also packages the ML model and related dependencies into a deployable artifact as a Docker image.
%\item  \textbf{New training data}.
%After data is mixed and cleaned, this updated training dataset is stored for use in retraining the ML model.
\item \textbf{CD}. 
A step where stakeholders automatically review the model's performance in development and staging environments, testing it with real-world scenarios to ensure stability before deployment to production. 
\end{itemize}
%Jenkins \cite{smart2011jenkins} is used to ensure a streamlined and efficient delivery process. Docker is used to containerize the ML model, while Kubernetes orchestrates its deployment, enabling seamless integration with cloud platforms such as AWS or GCP, along with automatic horizontal scalability management. For this work, the process concludes at the generation of the Docker image for deployment. Tools like Terraform and Ansible could further extend the pipeline by managing the deployment and configuration of cloud resources, including instances and load balancers, following an infrastructure-as-code paradigm. A theoretical comparison of regular retraining vs. data drift-triggered retraining is shown in Table \ref{tab:expected_outcomes}.
\subsection{Mathematics}
First, the KS test measures the difference between two empirical cumulative distribution functions. Given two distributions, $P(x)$ and $Q(x)$, the KS test is:  
\begin{equation}
  D_{KS} = \sup_x | F_P(x) - F_Q(x) |\,,  
\end{equation}
where $F_P(x)$ and $F_Q(x)$ are the empirical cumulative distribution functions of the original and new data distributions, respectively. A large $D_{KS}$ value indicates a significant shift.  

Second, the KL divergence quantifies how much one probability distribution $P(x)$ diverges from another reference distribution $Q(x)$:  
\begin{equation}
   D_{KL}(P || Q) = \sum_{x \in X} P(x) \log \frac{P(x)}{Q(x)}\,. 
\end{equation}
A higher $D_{KL}$ suggests greater divergence between  distributions.  

Third, 
 PSI is often used in monitoring changes in model input distributions. It is given by:  
\begin{equation}
    PSI = \sum_{i=1}^{n} (P_i - Q_i) \ln \left( \frac{P_i}{Q_i} \right)\,,
\end{equation}
where $P_i$ and $Q_i$ are the proportions of samples in bin $i$ for the reference and new distributions. A PSI value above $0.25$ typically indicates significant drift.  
Fourthly,
the MMD is a kernel-based method to compare two distributions:  
\begin{equation}
  MMD^2(P, Q) = \mathbb{E}_{x, x'} [ k(x, x') ] + \mathbb{E}_{y, y'} [ k(y, y') ] - 2 \mathbb{E}_{x, y} [ k(x, y) ]\,,  
\end{equation}
where $k(x, y)$ is a kernel function (e.g., Gaussian kernel). A large $MMD^2$ suggests distribution shift.

Finally, changes in model performance can also indicate data drift. The accuracy and F$_1$-score shifts are given by, respectively:  
\begin{equation}
    \Delta Acc = Acc_{new} - Acc_{old}\text { and }  \Delta F1 = F1_{new} - F1_{old}\,.
\end{equation}
A significant decrease in model performance suggests concept drift in the dataset.  

To obtain a single score, we use a weighted combination of all the above metrics:  
\begin{equation}
    DS = w_1 D_{KS} + w_2 D_{KL} + w_3 PSI + w_4 MMD + w_5 |\Delta Acc| + w_6 |\Delta F1|\,,
\end{equation}
where $w_1, w_2, w_3, w_4, w_5, w_6$ are hyperparameters that control the relative importance of each metric. These weights are set based on empirical studies.  
A threshold $\tau$ can be set such that if $DS > \tau$, retraining is triggered. 

\section{Experimental investigation}
For evaluation of our pipeline (Auto-MLOps), we conducted experiments using benchamrk datasets to simulate data distribution shifts  whose 
characteristics  are summarized in Table \ref{tab:dataset}.  
We evaluated the effectiveness of our approach using an auto-encoder (AE) \cite{masakuna2024streamlined}. The hyperparameters for AE are as follows: a weight decay of $10^{-6}$, batch sizes of 64, $100$ epochs, patience of 5 for early stop, learning rates will be set to $10^{-3}$ and Adam as optimizer.

 \begin{table}
	\centering
	\scalebox{.9}{
		\begin{tabular}{|c|c|c|c|c|}
			\hline
			\textbf{Data set} & \textbf{\# samples} & \textbf{\# attributes} & \textbf{anomaly (\%)}\\
			\hline
         	CICIOT \cite{neto2023ciciot2023}  & $416985$ &$41$ & $45$\\
			\hline
         	CREDIT \cite{warghade2020credit} & $234333$ &$30$ & $0.2$\\
			\hline
            ECG  \cite{khan2021ecg}  & $4998$ &$140$ & $58$\\
			\hline
      		IDS \cite{sharafaldin2018toward} & $430256$ &$95$ & $43$\\
            \hline
   			KITSUNE \cite{mirsky2018kitsune} & $210171$ &$116$ & $23$\\
			\hline
            MVTec \cite{bergmann2019mvtec}  & $5354$ &$1048576$ & $58$\\
			\hline
      		Visa \cite{zou2022spot} & $10821$ &$95$ & $11$\\
			\hline
	\end{tabular}}
	\caption{The characteristics of data sets. %Recall that training data contains normal datum only.
 }
	\label{tab:dataset}

\end{table}
We compared (Auto-MLOps) against three alternative retraining strategies: (1) a static model approach, where no retraining occurred regardless of data drift (STATIC); (2) a fixed-period retraining approach, where the model was retrained at predefined time intervals (e.g., daily, weekly, monthly) without considering data drift (FIXED); and (3) a drift-based naïve retraining
approach, which triggered retraining whenever a shift was detected but without
optimization for cost-effectiveness (NAIVE). 

\subsection{Results and discussion}
\begin{figure}[h]
	\centering
	\includegraphics[width=.8\textwidth]{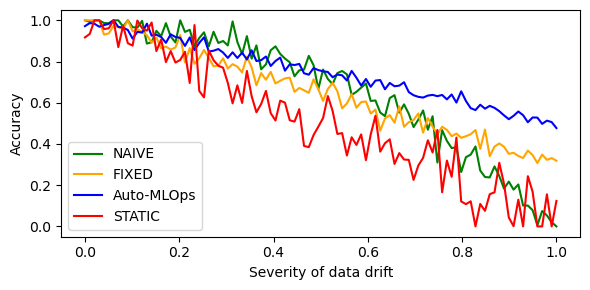}
	\caption{Accuracy degradation with increasing data drift severity.}
	\label{fig:degradation}
\end{figure}

\begin{figure}[h]
	\centering
	\includegraphics[width=1.\textwidth]{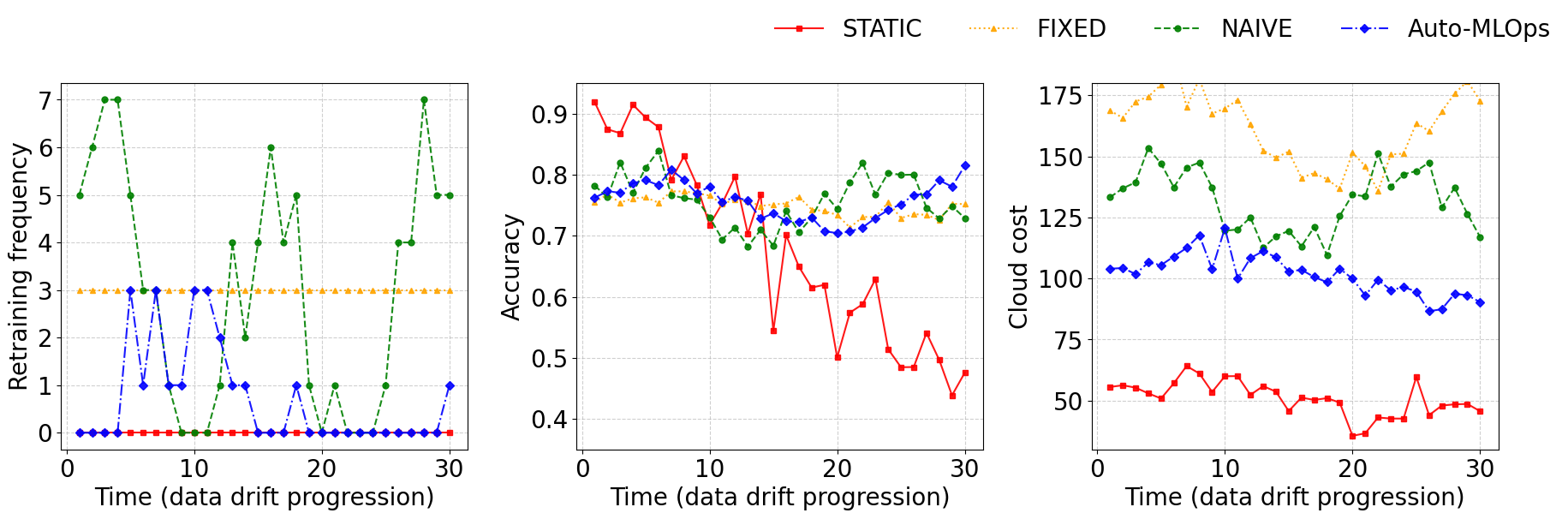}
	\caption{Average retraining frequency, accuracy and cloud cost across data sets.}
	\label{fig:result}
\end{figure}

Figure \ref{fig:degradation} illustrates the degradation in model accuracy as the severity of data drift increases. Figure \ref{fig:result} illustrates the retraining frequency, accuracy and cloud cost for each approach over time. Table \ref{tab:result} shows the average and standard deviation of performance of models across data sets.

As shown, all four models experience a decline in performance with rising data drift, but the extent of this degradation varies among the models. Our approach (Auto-MLOps) demonstrates the highest level of stability, maintaining a relatively consistent performance even as data drift intensifies. In comparison, FIXED, NAIVE and STATIC models exhibit a more pronounced drop in accuracy, highlighting their sensitivity to data shifts.

Figure \ref{fig:result} indicates that our proposed approach and NAIVE maintain a good accuracy while optimizing retraining costs. The STATIC model experienced a steady decline in accuracy over time due to increasing data drift. The FIXED approach partially mitigated this decline but incurred excessive cloud costs, as retraining was often performed unnecessarily when no significant drift had occurred. In contrast, our adaptive approach maintained model accuracy above 70\% while reducing retraining costs. 
This resulted in an optimal retraining frequency that balanced cost and model performance. Notably, during periods of minor distribution shifts, our approach deferred retraining without significant performance degradation, demonstrating its robustness in handling mild drift scenarios.

 \begin{table}
	\centering
	\scalebox{.9}{
		\begin{tabular}{|c|c|c|c|c|}
			\hline
			\textbf{Models} & \textbf{accuracy} & \textbf{cost} & \textbf{retraining frequency}\\
			\hline
         	STATIC   & $0.69\pm 0.2$ &$54.8\pm 5.6$ & $0\pm 0$\\
            FIXED   & $0.75\pm 0.02$ &$160.6\pm 10.5$ & $3\pm 0$\\
            NAIVE   & $0.75\pm 0.03$ &$130.5\pm 14.9$ & $4.3\pm 1.9$\\
            Auto-MLOps  & $0.75\pm 0.03$ &$108.1\pm 14.3$ & $1.4\pm 1.2$\\
			\hline
	\end{tabular}}
	\caption{Average and standard deviation of performance of models.
 }
	\label{tab:result}
\end{table}

The Auto-MLOps strategy demonstrates superior performance in managing a model lifecycle affected by data drift, achieving the highest accuracy ($0.75 \pm 0.03$) alongside the FIXED and NAIVE methods (Table \ref{tab:result}), but at a significantly reduced operational cost. The observed decay in the STATIC model's accuracy ($0.69 \pm 0.2$) highlights the presence of data drift. However, Auto-MLOps effectively mitigates this by intelligently determining the need for retraining, resulting in the lowest retraining frequency ($1.4 \pm 1.2$) among dynamic strategies and the lowest associated cost ($108.1 \pm 14.3$), thereby confirming that a drift-aware, automated pipeline provides the most efficient balance between model performance preservation and computational resource management.

Our findings highlight the importance of cost-aware, adaptive retraining in cloud-based MLOps pipelines. Unlike conventional approaches that either neglect retraining or rely on rigid scheduling, our method dynamically adapts to data shifts while minimizing redundant computation. This is beneficial for applications with budget constraints.

%Overall, our results demonstrate that a multi-criteria optimization approach to automated retraining provides a significant advantage over traditional retraining methods, achieving both model robustness and cost efficiency. By dynamically adapting to data shifts and optimizing retraining decisions, our MLOps pipeline contributes to the advancement of sustainable and cost-effective machine learning deployment strategies in cloud environments.
\label{sec:results}

\section{Conclusion}
\label{sec:conclusion}
Automating MLOps pipeline that combines different statistical metrics to detect significant changes in data distribution is an innovative contribution. 
This is not just a technical improvement; it directly translates into cost savings and operational efficiency. Companies leveraging cloud services for ML workloads already face significant expenses. If retraining is  justified by significant data changes, companies can avoid redundant computation and reduce their annual cloud expenditure, ultimately  making their MLOps more sustainable and scalable.

Despite its advantages, our approach has certain limitations. The effectiveness of drift detection depends on the sensitivity of the statistical methods used, and false positives could still lead to unnecessary retraining. Future work can explore reinforcement learning-based retraining policies, enabling the pipeline to learn optimal retraining schedules from past deployment data. Also, integrating more sophisticated cost models that incorporate real-time cloud pricing fluctuations could enhance cost efficiency even further.
\bibliographystyle{splncs04}
%\section*{Declarations}
%\subsection*{Ethical Approval}
%Not applicable.
 
%\subsection*{Competing interests }
%The authors declare that they have no competing interests. 
%\subsection*{Authors' contributions }
%Authors have proposed an automated MLOps pipeline for  retraining a ML model only when data distribution have changed significantly. Jordan F. Masakuna came up with the idea. David O. Dimandja ran experiments.
%\subsection*{Funding}

%Not applicable.
 
%\subsection*{Availability of data and materials} 
%Our code can be found at \url{https://github.com/jmf-mas/auto-mlops}. 
\bibliography{references}

@article{feurer2019hyperparameter,
  title={{Hyperparameter Optimization}},
  author={Feurer, Matthias and Hutter, Frank},
  journal={Automated machine learning: Methods, systems, challenges},
  pages={3--33},
  year={2019},
  publisher={Springer International Publishing}
}

@misc{berrar2019cross,
  title={{Cross-Validation.}},
  author={Berrar, Daniel and others},
  year={2019}
}

@article{kreuzberger2023machine,
  title={{Machine Learning Operations (MLOps): Overview, Definition, and Architecture}},
  author={Kreuzberger, Dominik and K{\"u}hl, Niklas and Hirschl, Sebastian},
  journal={IEEE Access},
  volume={11},
  pages={31866--31879},
  year={2023},
  publisher={IEEE}
}

@book{sterling2017high,
  title={{High Performance Computing: Modern Systems and Practices}},
  author={Sterling, Thomas and Brodowicz, Maciej and Anderson, Matthew},
  year={2017},
  publisher={Morgan Kaufmann}
}

@article{lu2018learning,
  title={{Learning under Concept Drift: A Review}},
  author={Lu, Jie and Liu, Anjin and Dong, Fan and Gu, Feng and Gama, Joao and Zhang, Guangquan},
  journal={IEEE transactions on knowledge and data engineering},
  volume={31},
  number={12},
  pages={2346--2363},
  year={2018},
  publisher={IEEE}
}

@article{dongare2012introduction,
  title={{Introduction to Artificial Neural Network}},
  author={Dongare, AD and Kharde, RR and Kachare, Amit D and others},
  journal={International Journal of Engineering and Innovative Technology (IJEIT)},
  volume={2},
  number={1},
  pages={189--194},
  year={2012},
  publisher={Citeseer}
}

@article{berger2014kolmogorov,
  title={{Kolmogorov--Smirnov Test: Overview}},
  author={Berger, Vance W and Zhou, YanYan},
  journal={Wiley statsref: Statistics reference online},
  year={2014},
  publisher={Wiley Online Library}
}

@article{ji2020kullback,
  title={{Kullback--Leibler Divergence Metric Learning}},
  author={Ji, Shuyi and Zhang, Zizhao and Ying, Shihui and Wang, Liejun and Zhao, Xibin and Gao, Yue},
  journal={IEEE transactions on cybernetics},
  volume={52},
  number={4},
  pages={2047--2058},
  year={2020},
  publisher={IEEE}
}

@book{yurdakul2018statistical,
  title={{Statistical Properties of Population Stability Index}},
  author={Yurdakul, Bilal},
  year={2018},
  publisher={Western Michigan University}
}

@inproceedings{smola2006maximum,
  title={{Maximum Mean Discrepancy}},
  author={Smola, Alexander J and Gretton, A and Borgwardt, K},
  booktitle={13th international conference, ICONIP},
  pages={3--6},
  year={2006}
}

@article{masakuna2023prior,
  title={{Do Prior Information on Performance of Individual Classifiers for Fusion of Probabilistic Classifier Outputs Matter?}},
  author={Masakuna, Jordan Felicien and Kafunda, Pierre Katalay},
  journal={Journal of Classification},
  volume={40},
  number={3},
  pages={468--487},
  year={2023},
  publisher={Springer}
}

@article{mahadevan2024cost,
  title={{Cost-aware Retraining for Machine Learning}},
  author={Mahadevan, Ananth and Mathioudakis, Michael},
  journal={Knowledge-Based Systems},
  volume={293},
  pages={111610},
  year={2024},
  publisher={Elsevier}
}

@article{ditzler2015learning,
  title={{Learning in Nonstationary Environments: A Survey}},
  author={Ditzler, Gregory and Roveri, Manuel and Alippi, Cesare and Polikar, Robi},
  journal={IEEE Computational Intelligence Magazine},
  volume={10},
  number={4},
  pages={12--25},
  year={2015},
  publisher={IEEE}
}

@article{sculley2015hidden,
  title={{Hidden Technical Debt in Machine Learning Systems}},
  author={Sculley, David and Holt, Gary and Golovin, Daniel and Davydov, Eugene and Phillips, Todd and Ebner, Dietmar and Chaudhary, Vinay and Young, Michael and Crespo, Jean-Francois and Dennison, Dan},
  journal={Advances in neural information processing systems},
  volume={28},
  year={2015}
}

@article{louppe2017learning,
  title={{Learning to Pivot with Adversarial Networks}},
  author={Louppe, Gilles and Kagan, Michael and Cranmer, Kyle},
  journal={Advances in neural information processing systems},
  volume={30},
  year={2017}
}

@phdthesis{masakuna2020active,
  title={{Active Strategies for Coordination of Solitary Robots}},
  author={Masakuna, Jordan Felicien},
  year={2020},
  school={Stellenbosch: Stellenbosch University}
}

@inproceedings{bifet2007learning,
  title={{Learning from Time-Changing Data with Adaptive Windowing}},
  author={Bifet, Albert and Gavalda, Ricard},
  booktitle={Proceedings of the 2007 SIAM international conference on data mining},
  pages={443--448},
  year={2007},
  organization={SIAM}
}

@article{bayram2022concept,
  title={{From Concept Drift to Model Degradation: An Overview on Performance-Aware Drift Detectors}},
  author={Bayram, Firas and Ahmed, Bestoun S and Kassler, Andreas},
  journal={Knowledge-Based Systems},
  volume={245},
  pages={108632},
  year={2022},
  publisher={Elsevier}
}

@inproceedings{zou2022spot,
  title={{Spot-the-Difference Self-Supervised Pre-Training for Anomaly Detection and Segmentation}},
  author={Zou, Yang and Jeong, Jongheon and Pemula, Latha and Zhang, Dongqing and Dabeer, Onkar},
  booktitle={European Conference on Computer Vision},
  pages={392--408},
  year={2022},
  organization={Springer}
}

@inproceedings{bergmann2019mvtec,
  title={{A Comprehensive Real-World Dataset for Unsupervised Anomaly Detection}},
  author={Bergmann, Paul and Fauser, Michael and Sattlegger, David and Steger, Carsten},
  booktitle={Proceedings of the IEEE/CVF Conference on Computer Vision and Pattern Recognition},
  pages={9592--9600},
  year={2019}
}

@article{khan2021ecg,
  title={{ECG Images Dataset of Cardiac and COVID-19 Patients}},
  author={Khan, Ali Haider and Hussain, Muzammil and Malik, Muhammad Kamran},
  journal={Data in Brief},
  volume={34},
  pages={106762},
  year={2021},
  publisher={Elsevier}
}

@article{warghade2020credit,
  title={{Credit Card Fraud Detection from Imbalanced Dataset using Machine Learning Algorithm}},
  author={Warghade, Swati and Desai, Shubhada and Patil, Vijay},
  journal={International Journal of Computer Trends and Technology},
  volume={68},
  number={3},
  pages={22--28},
  year={2020}
}

@article{sharafaldin2018toward,
  title={{Toward Generating a New Intrusion Detection Dataset and Intrusion Traffic Characterization}},
  author={Sharafaldin, Iman and Lashkari, Arash Habibi and Ghorbani, Ali A},
  year={2018},
journal={International Conference on Information Systems Security and Privacy}
}

@inproceedings{mirsky2018kitsune, 
title={{Kitsune: An Ensemble of Autoencoders for Online Network Intrusion Detection}}, 
author={Mirsky, Yisroel and Doitshman, Tomer and Elovici, Yuval and Shabtai, Asaf}, booktitle={Network and Distributed System Security Symposium (NDSS)}, 
year={2018} }

@article{neto2023ciciot2023,
  title={{CICIoT2023: A Real-Time Dataset and Benchmark for Large-Scale Attacks in IoT Environment}},
  author={Neto, Euclides Carlos Pinto and Dadkhah, Sajjad and Ferreira, Raphael and Zohourian, Alireza and Lu, Rongxing and Ghorbani, Ali A},
  year={2023},
  journal={Preprints}
}

@article{masakuna2024streamlined,
  title={{Streamlined and Resource-Efficient Estimation of
Epistemic Uncertainty in Deep Ensemble
Classification Decision via Regression}},
  author={Masakuna, Jordan F and D'Jeff, K Nkashama and Soltani, Arian and Frappier, Marc and Tardif, Pierre-Martin and Kabanza, Froduald},
  journal={IEEE Transactions on Emerging Topics in Computational Intelligence},
  year={2024},
  publisher={IEEE}
}

%\bibliographystyle{splncs04}
%\bibliography{mybibliography}
%

\end{document}